\documentclass{article}
\usepackage{spconf,amsmath,epsfig}

\usepackage{cite}
\usepackage{amssymb,amsfonts}
\usepackage{algorithmic}
\usepackage{graphicx}
\usepackage{textcomp}
\usepackage{xcolor}
\usepackage{tabularx,booktabs}
\usepackage{hyperref}
\usepackage{siunitx}
\usepackage{float}

\title{Optimal Transport for Change Detection on LiDAR Point Clouds}
\name{Marco Fiorucci\textsuperscript{1,3}, Peter Naylor\textsuperscript{2}, Makoto Yamada\textsuperscript{2,3,4}}

\address{\textsuperscript{1}Center for Cultural Heritage Technology, Istituto Italiano di Tecnologia, Venice, Italy \\ \textsuperscript{2}RIKEN AIP, Kyoto, Japan,  \textsuperscript{3}Kyoto University, Kyoto, Japan, \\ \textsuperscript{4}Okinawa Institute of Science and Technology, Okinawa, Japan}

\begin{document}
%
\maketitle
\begin{abstract}
Unsupervised change detection between airborne LiDAR data points, taken at separate times over the same location, can be difficult due to unmatching spatial support and noise from the acquisition system.
Most current approaches to detect changes in point clouds rely heavily on the computation of Digital Elevation Models (DEM) images and supervised methods. 
Obtaining a DEM leads to LiDAR informational loss due to pixelisation, and supervision requires large amounts of labelled data often unavailable in real-world scenarios.
We propose an unsupervised approach based on the computation of the transport of 3D LiDAR points over two temporal supports.
The method is based on unbalanced optimal transport and can be generalised to any change detection problem with LiDAR data.
We apply our approach to publicly available datasets for monitoring urban sprawling in various noise and resolution configurations that mimic several sensors used in practice. 
Our method allows for unsupervised multi-class classification and outperforms the previous state-of-the-art unsupervised approaches by a significant margin.
\end{abstract}
\section{Introduction}
\label{sec:intro}
Detecting changes occurring in multi-temporal remote sensing data plays a crucial role in monitoring several aspects of real life, such as disasters, deforestation, and urban planning \cite{Asokan2019ESI, Shi2020_RS}. 
In the latter context, identifying both newly built and demolished buildings is essential to help landscape and city managers to promote sustainable development \cite{Reynolds2017_RS, Erdogan2019_IJRS}. 
While the use of airborne LiDAR point clouds has become widespread in urban change detection \cite{deGelis2021_RS, STILLA2023}, the most common approaches require the transformation of a point cloud into a regular grid of interpolated height measurements, i.e. Digital Elevation Model (DEM) \cite{Tuong2004IGARSS, ZHOU2020JPRS, Lyu2020IJGI}.     
However, the DEM's interpolation step causes an information loss related to the height of the objects, affecting the detection capability of building changes, where the high resolution of LiDAR point clouds in the third dimension would be the most beneficial.

Notwithstanding recent attempts to detect changes directly on 3D point clouds using either a semantic segmentation processing step \cite{xu2015RS, dai2020RS} or developing a novel Siamese network based on a Kernel Point Convolution (KPC) \cite{2023DEGELIS}, all of which supervised,  hence require a fully labelled training dataset. However, the creation of a training dataset for change detection on a bi-temporal pair of 3D point clouds requires manually labelling millions of points to take into account both the difference in the spatial resolution (i.e., the number of points per square meter) and the lack of registration between the two point clouds of each pair. 

 \begin{figure*}[ht]
\includegraphics[width=\linewidth]{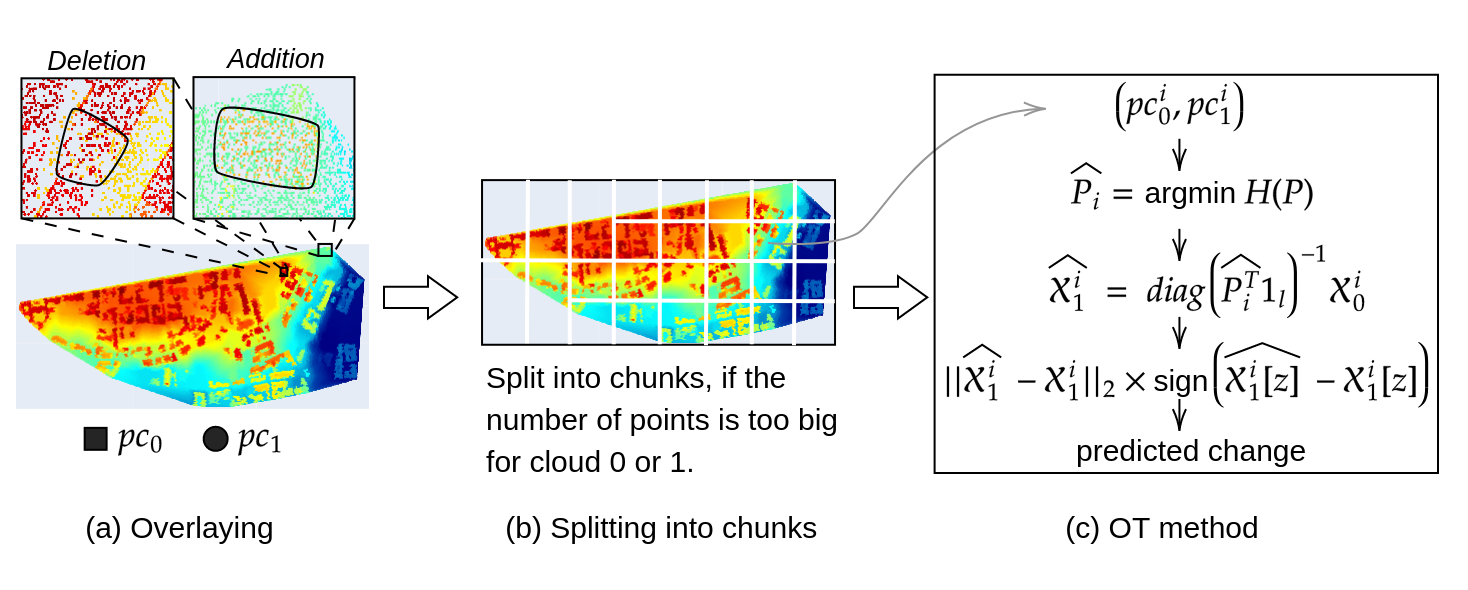} \vspace{-1cm}
\caption{Method pipeline with chunk splitting and OT change detection applied to chunk $i$.} \vspace{-0.5cm}
\label{fig:method}
\end{figure*} 

These limitations call for developing unsupervised methods for identifying change directly on 3D point clouds. To the best of our knowledge, based on the survey papers \cite{deGelis2021_RS, 2022Kharroubi}, there exists only one unsupervised approach able to identify both positive and negative changes (new and demolished buildings, respectively) directly on 3D point clouds, namely the Multiscale Model to model Cloud Comparison (M3C2) \cite{lague2013ISPRS}, which is a distance-based change detection method. 
Specifically, each point cloud of a bi-temporal pair is stored in an octree structure, a recursive subdivision of the 3D space, to ensure a fast identification of the nearest neighbours of each point. The identification of changes is performed by thresholding the Hausdorff distance between each point of the first cloud and each neighbour's points of the second cloud (i.e., all the points contained in the corresponding homologous cell of the octree) laying on a local normal surface. 
Despite the M3C2 being capable of quickly processing bi-temporal pairs with millions of points, it is sensitive to the direction and amplitude of changes \cite{2022Kharroubi}, which in turn affect the change detection performance.  

Motivated by the previous arguments, we introduce a principled change detection pipeline based on Optimal Transport (OT), capable of distinguishing between newly built and demolished buildings. OT theory provides us with a methodology to project the first point cloud on the support of the second one through the so-called displacement interpolation \cite{2015Solomon} by considering both clouds as two uniform probability distributions. The changes are then computed point-wise between the projected and the second cloud. OT theory does not require that the two clouds have the same point density and is insensitive to the direction and amplitude of changes. 
Courty et al. \cite{Courty2016IGARSS} proposed to use OT to identify changes occurred on a coastal cliff in a pair of LiDAR point clouds. However, their algorithm requires the conservation of mass between the two point clouds, which models them as two probability distributions and detects only change, irrespective of it being a positive or negative change. 
We argue that this shortcoming affects the change detection performance. To address this limitation, we propose to use unbalanced optimal transport \cite{2020Pham} to cope with the creation and destruction of mass related to building changes occurring on a bi-temporal pair of LiDAR point clouds.
We demonstrate the efficacy of our approach on the only publicly available airborne LiDAR dataset for change detection \cite{deGelis2021_RS} by showing superior performance over the M3C2 and the OT-based method introduced in \cite{Courty2016IGARSS}.

\section{Methodology}

We aim to detect the changes occurring on a bi-temporal pair of airborne LiDAR point clouds, namely $pc_0$ and $pc_1$ collected at time $t_0$ and $t_1 > t_0$ over the same geographical area. 
In this setting, detecting change is solved by projecting the support of $\mathcal{X}_0 \in \mathbb{R}^3$ onto the support of $\mathcal{X}_1 \in \mathbb{R}^3$, i.e., matching the probability distributions $\mu_0$ and $\mu_1$ modelling $pc_0$ and $pc_1$ respectively. 
After this projection step, changes are computed as the pointwise distance between the points of the projected space, namely $\hat{\mathcal{X}_1}$, and the space $\mathcal{X}_1$.  

An effective way to match two probability distributions is addressed by Optimal Transport (OT) theory \cite{villani2008OT, santambrogio2015OT, Perye2018COT} that enables to transport $\mathcal{X}_0$ onto $\mathcal{X}_1$ following a least effort principle, i.e. a transport called OT plan, that is optimal given a cost metric $C$ on $(\mathcal{X}_0 \times \mathcal{X}_1$). 
The problem of computing the OT plan $\hat P$ is defined as $\hat P = \arg \min_{P \in \Pi}  H(P)$  s.t. $ H(P) = <P,C> + \lambda R(P)$ 
and $\Pi = \{ P \in \mathbb{R}^{n_0 \times n_1} | \, P1_{n_1} = \mu_0, P^T 1_{n_0} = \mu_1 \}$, where $1_l$ is the $1$-dimensional vector of ones, $<>$ is the Frobenius dot product, $R(P)$ is the
entropic regularisation term and $\lambda$ the associated scalar. To detect a change in point cloud pairs, Courty et al. \cite{Courty2016IGARSS} proposed to compute the projection of $\mathcal{X}_0$ onto $\mathcal{X}_1$ using the barycenter projection \cite{villani2008OT} defined as $\hat{\mathcal{X}_1} = diag(\hat P^T 1_{n_0})^{-1} \mathcal{X}_0$, where the OT plan $\hat P$ minimises $H$.  

We advocate that the use of unbalanced OT \cite{2020Pham} will increase the performance because it is more effective to cope with the creation and destruction of mass related to building changes than the previous version \cite{Courty2016IGARSS}. 
The unbalanced version adds another regularisation term to $H$: $R_u (P) = KL(P 1_{n_1}, [1 / n_0]_{n_0})$ that penalises mass conservation, which is not valid when the matching between the two distributions $\mu_1,\mu_2$ (modelling $pc_0,pc_1$) is not surjective \cite{Sejourne2019, Chapel2021UnbalancedOT} as in the case of urban changes. 


\begin{table}[ht]
\centering
\begin{tabular}{c|c|c|c}
\toprule
Dataset & M3C2   &  OT  & UOT \\
\midrule
\midrule
Low res - low noise   & 0.2987 & 0.2702 & 0.4065          \\
High res - low noise    & 0.5373 & 0.3421 & 0.5520          \\
Low res - high noise   & 0.3872 & 0.2694 & 0.3926          \\
Photogrammetry  & 0.3501 & 0.2849 & 0.3989          \\
MultiSensor  & 0.3778 & 0.3036 & 0.4817         \\
\bottomrule
\end{tabular}
\caption{Intersection over union benchmark with state-of-the-art on the five datasets for building change detection.} \vspace{-0.5cm}
\label{tbl:res}
\end{table}

\section{Experimental Results}

\begin{figure*}[ht]
\centering
\includegraphics[width=0.8\linewidth]{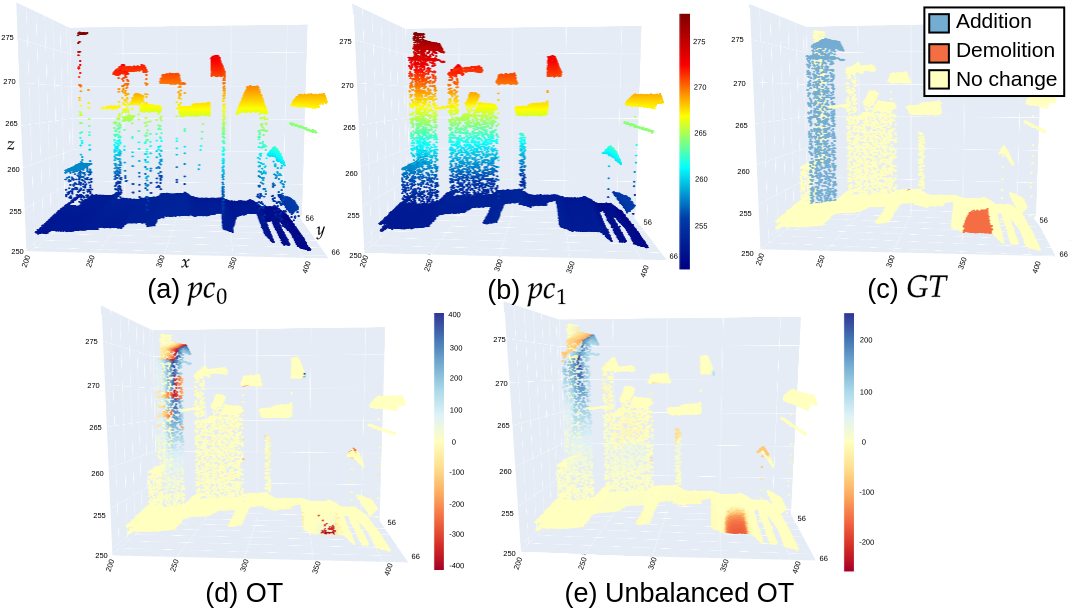}
\caption{Comparison between OT \cite{Courty2016IGARSS} and our method on two point clouds ($pc_0$, $pc_1$) with a different point density. Unbalanced OT detected changes which are not properly identified by OT, such as the demolished building (bottom right of the GT).}
\label{fig:example}
\end{figure*}

In this section, we evaluate our change detection pipeline on the only publicly available dataset for building change detection \cite{deGelis2021_RS}, which contains bi-temporal pair of airborne LiDAR point clouds. 
This dataset is composed of five sub-datasets characterised by different spatial resolution and acquisition noise levels: from high resolution with low noise to low resolution with noisy point clouds. 
Fig. \ref{fig:method} shows our pipeline, composed of the pre-processing step, which divides each point cloud into chunks, and the change detection module that first computes the barycenter projection using unbalanced OT and, subsequently, the pointwise distance. Dividing the point cloud into chunks is necessary because they usually contain millions of points which are too large for OT computation due to the quadratic complexity \cite{Perye2018COT}.
We compared the unbalanced OT with both the previous OT version \cite{Courty2016IGARSS} and the M3C2 method \cite{lague2013ISPRS} on the five sub-datasets. 
Following the experimental setting presented in \cite{deGelis2021_RS}, we used empirical thresholds for these three methods to identify points in $pc_1$  that changed with respect to $pc_0$ by comparing if the pointwise distance exceeds the threshold.
The threshold is computed by trying several values and systematically selecting the best one for each sub-dataset as this is reported in similar studies for a fair comparison \cite{deGelis2021_RS}. 
Table \ref{tbl:res} shows that our pipeline, using unbalanced OT, outperforms the other two state-of-the-art methods and is robust to the resolution level and input noise. In particular, we demonstrated experimentally (quantitatively in Table \ref{tbl:res}, and qualitatively in Figure \ref{fig:example}) our related hypothesis, which is that unbalanced optimal transport handles the creation and destruction of mass related to building changes more efficiently, while the (balanced) OT algorithm \cite{Courty2016IGARSS} is less effective in all the five sub-datasets.

Computationally speaking, a cloud containing millions of points is too large for optimal transport, which has quadratic time complexity.
To compute the optimal transport plan for each chunk on a $16$ GB GPU, we fix the upper limit for a cloud point to $30\,000$ points.
For example, this requirement will slit the largest dataset, i.e., high resolution and low noise, into $162$ chunks.
As the data can be noisy and non-uniformly distributed over the spatial dimensions, we cannot ensure that each chunk has the same number of points. 
The chunks computed from the largest dataset range from a minimum size of $8\,072$ points to a maximum of $29\,346$ with a median value of $24\,276$ points.
The memory footprint on a GPU ranges from $3.3$ to $9.7$ GB with a median of $8.8 $ GB, and the computation time ranges from $4.8$ to $10.7$ seconds with a median of $7.94$ seconds.
The entire pipeline is fully reproducible on HPC\footnote{\url{https://github.com/MarcoFiorucci/OT-4-change-detection}}.


\section{Conclusions}
We introduced an unsupervised change detection pipeline based on unbalanced OT, which provides us with a general method to identify changes directly on LiDAR point clouds.
We successfully validated our framework showing that it surpasses two state-of-the-art methods for identifying both newly built and demolished buildings. 
We believe that OT can effectively address different problems of interest in the remote sensing community, such as flood and fire detection, climate changes and cultural heritage. 
This work paves the way for a principled method for change detection on different remote sensing non-registered time series by combining OT theory with image processing and machine learning. 

\subsection*{Acknowledgement}
This project has received funding from the European Union’s Horizon 2020 research and innovation programme under grant agreement No 101027956.

\bibliographystyle{IEEEbib}
\bibliography{refs}

\end{document}